\documentclass{sig-alternate-05-2015}
\usepackage{hyperref}
\usepackage{algorithm}
\usepackage{algpseudocode}
\usepackage{amsmath}
\usepackage{amssymb}
\usepackage[utf8]{inputenc} % allow utf-8 input
\usepackage[T1]{fontenc}    % use 8-bit T1 fonts
\usepackage{hyperref}       % hyperlinks
\usepackage{url}            % simple URL typesetting
\usepackage{booktabs}       % professional-quality tables
\usepackage{amsfonts,amsmath,amssymb}       % blackboard math symbols
\usepackage{nicefrac}       % compact symbols for 1/2, etc.
\usepackage{microtype}      % microtypography
\usepackage{graphicx}

\begin{document}

% Copyright
\setcopyright{acmcopyright}
%\setcopyright{acmlicensed}
%\setcopyright{rightsretained}
%\setcopyright{usgov}
%\setcopyright{usgovmixed}
%\setcopyright{cagov}
%\setcopyright{cagovmixed}

% DOI
%\doi{10.475/123_4}

% ISBN
%\isbn{123-4567-24-567/08/06}

%Conference
%\conferenceinfo{PLDI '13}{June 16--19, 2013, Seattle, WA, USA}

%\acmPrice{\$15.00}

%
% --- Author Metadata here ---
%\conferenceinfo{WOODSTOCK}{'97 El Paso, Texas USA}
%\CopyrightYear{2007} % Allows default copyright year (20XX) to be over-ridden - IF NEED BE.
%\crdata{0-12345-67-8/90/01}  % Allows default copyright data (0-89791-88-6/97/05) to be over-ridden - IF NEED BE.
% --- End of Author Metadata ---

\title{Outlier Detection by Consistent Data Selection Method}
%\subtitle{[Extended Abstract]
%\titlenote{A full version of this paper is available as
%\textit{Author's Guide to Preparing ACM SIG Proceedings Using
%\LaTeX$2_\epsilon$\ and BibTeX} at

\numberofauthors{2} %  in this sample file, there are a *total*
\author{
% 1st. author
\alignauthor
Utkarsh Porwal\\
       \affaddr{eBay Inc}\\
       \affaddr{San Jose, CA}\\
      % \affaddr{Wallamaloo, New Zealand}\\
       \email{uporwal@ebay.com}
% 2nd. author
\alignauthor
Smruthi Mukund\titlenote{This work was done while the author worked at eBay}\\
       \affaddr{eBay Inc}\\
       \affaddr{San Jose, CA}\\
       %\affaddr{Dublin, Ohio 43017-6221}\\
       \email{smukund@buffalo.edu}
}
% There's nothing stopping you putting the seventh, eighth, etc.
% author on the opening page (as the 'third row') but we ask,
% for aesthetic reasons that you place these 'additional authors'
% in the \additional authors block, viz.
% Just remember to make sure that the TOTAL number of authors
% is the number that will appear on the first page PLUS the
% number that will appear in the \additionalauthors section.

\maketitle
\begin{abstract}
Often the challenge associated with tasks like fraud and spam detection\cite{acuna2004meta} is the lack of all likely patterns needed to train suitable supervised learning models. In order to overcome this limitation, such tasks are attempted as outlier or anomaly detection tasks. We also hypothesize that outliers have behavioral patterns that change over time. Limited data and continuously changing patterns makes learning significantly difficult. In this work we are proposing an approach that detects outliers in large data sets by relying on data points that are consistent. The primary contribution of this work is that it will quickly help retrieve samples for both consistent and non-outlier data sets and is also mindful of new outlier patterns. No prior knowledge of each set is required to extract the samples. The method consists of two phases, in the first phase, consistent data points (non-outliers) are retrieved by an ensemble method of unsupervised clustering techniques and in the second phase a one class classifier trained on the consistent data point set is applied on the remaining sample set to identify the outliers. The approach is tested on three publicly available data sets and the performance scores are competitive.
\end{abstract}

%\begin{CCSXML}
%<ccs2012>
%  <concept_id>10010520.10010553.10010562</concept_id>
%  <concept_desc>Computer systems organization~Embedded systems</concept_desc>
%  <concept_significance>500</concept_significance>
%  <concept_id>10010520.10010575.10010755</concept_id>
%  <concept_desc>Computer systems organization~Redundancy</concept_desc>
%  <concept_significance>300</concept_significance>
%  <concept_id>10010520.10010553.10010554</concept_id>
%  <concept_desc>Computer systems organization~Robotics</concept_desc>
%  <concept_significance>100</concept_significance>
%  <concept_id>10003033.10003083.10003095</concept_id>
%  <concept_desc>Networks~Network reliability</concept_desc>
%  <concept_significance>100</concept_significance>
%</ccs2012>  
%\end{CCSXML}
%
%\ccsdesc[500]{Computer systems organization~Embedded systems}


%
%  Use this command to print the description
%
\printccsdesc

% We no longer use \terms command

\keywords{Outlier Detection; Consistent Data Selection}

\section{Introduction}
Tasks like credit card fraud detection, e-Commerce fraud detection, voting irregularity analysis, severe weather predictions are attempted as outlier detection, intrusion detection or anomaly detection tasks\cite{hawkins1980identification}\cite{vic1994outliers}\cite{ruts1996computing}\cite{Fawcett97adaptivefraud}\cite{Johnson98fastcomputation}\cite{penny2001comparison}\cite{acuna2004meta}\cite{lu2003algorithms}. The goal in all of the above tasks is to find those data points that contain useful information on abnormal behavior of the system described by the data. In most cases, the ratio of the number of outlier data points to the number of consistent data points in the sample set is relatively small. This poses a challenge to supervised machine learning methods that rely on using training data sets\cite{Chawla}. Unsupervised techniques of clustering have shown to successfully identify outliers\cite{Loureiro04outlierdetection}. However, such techniques label outliers as noise as they do not belong to the dense pool of clusters. When the sample set is very large and the data characteristics to identify outliers span over a large number of dimensions, such techniques are unable to effectively capture the deviation. In order to address this problem several techniques that rely on subspace clustering and trajectory detection where different dense localities of the data are effectively identified based on the different subset of attributes used for clustering\cite{ramaswamy2000efficient}. These methods are complex and are not easy to implement in a distributed fashion for processing large data sets. Moreover, in all of these approaches identifying outliers has been the focus. Similarly, approaches that model on a single class - one class classifiers\cite{scholkopf2000new} would require samples that exhibit good behavior making prior knowledge of the good behavior essential in extracting the samples.
In this work, we formulate the problem as a consistent data point detection problem, where the goal is to effectively identify non-outlier (consistent) data points in the sample set. We do not seek to retrieve anomalies from the data. On the other hand our goal is to extract data that has lowest variance. In other words, extract data points that can perfectly represent good behavior or desired/normal pattern. Therefore, the fact that in high dimensional data, clusters may be present in different affine subspaces offers no hindrance in retrieving the most consistent data points. Our only requirement is that data points should be the part of the same cluster in all the subspaces pertaining to different set of subspaces. For the same reason projecting the data from higher manifold to a lower one will not hurt the performance of our system as our assumption is that the data is projected on to a higher manifold while it is actually embedded in a lower one. This assumption is a reasonable assumption to make in applications where many attributes have high correlation coefficients. Once we have successfully identified consistent data points, we apply a one-class classifier on this data set to determine the outliers.
The advantages of our method is that (i) no prior knowledge of either the outlier or the consistent sample set is required (ii) rapidly helps to bootstrap samples that represent the two classes that can further be analyzed, making quick discovery of new fraudulent patterns (iii) the algorithm suggested for clustering can easily be implemented in a distributed system there by not being inhibited by the size of the data set that needs to be processed\cite{Li:2011}.

Our first step to identifying the consistent data points is based on an ensemble of clustering techniques. Ensemble of learners is proven to be more effective than a single learner and is been used heavily in supervised setting\cite{dietterich2000ensemble}\cite{porwal2012ensemble}. However, ensemble of unsupervised methods is also been used for the task of outlier detection\cite{aggarwal2015theoretical}.As highlighted by Ana et al.\cite{Fred} clustering technique is effective in identifying consistent clusters in the sample set. In this work, we use the popular $k$-means clustering technique in an ensemble fashion and cluster data into different
$k$ values. A similarity measure is then used to measure consistent cluster associations for each datapoint.
The second step is to detect the outliers. Consistent data points found in the first step are used to train a one class classifier. The original application of one-class classifiers was in outlier detection\cite{shin2005one}. One-class classification is significant in applications where only data belonging to one class in a two class problem is easy to obtain. Data of the other class is either difficult to obtain or is expensive to collect. Tax et al.\cite{tax2003online} and Manevitz et al.\cite{Manevitz} have used support vector machines using only the consistent examples. The main idea here is to construct a decision boundary around the consistent data samples (positives) so as to differentiate the outliers from the positive samples. The advantage of this method is that it minimizes the need for ground truth data for both classes. In this work, a one class classifier trained on the consistent data point set to determine the outliers.
The performance of our approach is tested on three publicly available UCI Datasets\cite{Lichman:2013} - the ionosphere dataset, the arrhythmia dataset and the musk dataset. We show that our method is effective in identifying the consistent data points and helps determine the outlier points with high confidence.
The rest of the paper is organized as follows. Section~\ref{sec:related} describes the related research in the area of outlier detection. Section~\ref{sec:hypo} explains our hypothesis using a synthetic data set. Section~\ref{sec:cd} outlines the algorithm for consistent data set detection. Data sets descriptions and preparation is outlined in Section~\ref{sec:exp}. Results for each of the data sets are explained in the subsections of Section~\ref{sec:result}. Section~\ref{sec:observation} highlights the observations followed by conclusion in Section~\ref{sec:conclusion}.

\section{Related Work}
\label{sec:related}
Several approaches have been proposed for anomaly detection based on different assumptions and techniques. One of the most common and intuitive way to detect anomalies is by assuming an underlying distribution for the data\cite{horn2001effect}. Sufficient statistics (mean and standard deviation in case of Gaussian distribution) are calculated and data points are flagged as anomalies based on different statistical tests. One such example where sufficient statistics help is when trying to find data points lying in the tail of the distribution. There are different variant to this approach. One is by modeling all the attributes together in one multi-dimensional function and two, by modeling each attribute with a distribution function \cite{grubbs1969procedures}\cite{Laurikkala00informalidentification}. The latter can be considered as a special case of multi-dimensional approach as the non-diagonal elements of the covariance matrix are zero. Another approach is based on the distance measure. In this approach it is assumed that all regular data points lie close to each other and anomalies are far from them\cite{Tan:2005}. Nearest neighbor techniques have been employed to detect anomalies with this assumption\cite{ramaswamy2000efficient}. However, this approach works only for low dimensional data. In higher dimensions concept of distance loses its meaning rendering this approach inefficient\cite{aggarwal2001surprising}. This argument is also valid for density-based measures to detect anomalies. Therefore, distance and density-based approaches falls short in case of high dimensional data. Another approach is clustering based approach where it is assumed that regular data points make clusters and anomalies are either not part of any cluster or make separate clusters\cite{Tan:2005}\cite{Jain:1988}. However, for high dimensional data there are multiple challenges. In multi-dimensional data, data often make different clusters for different set of attributes and these clusters lie in different subspaces. Therefore, sub space clustering is performed with an assumption that all the subspaces are axis parallel to reduce the complexity of exploring subspaces. Although there is some work proposed for randomly aligned subspace clustering, it is to be noted that it is often not possible to reduce the dimensionality of the data and apply clustering in a lower manifold because of the \textit{local feature relevance}\cite{Kriegel:2009}.  Chandola et al. \cite{Chandola:2009} and  Kriegel et al.\cite{Kriegel:2009} covers a more comprehensive survey on related work in the field of anomaly detection and clustering in high dimensional data.

\section{Hypothesis}
\label{sec:hypo}
Consider an application of credit card fraud detection. Here, it is much easier for us to obtain samples with good non-fraud behavior than samples that exhibit a fraudulent pattern as the latter is time variant. Once a fraud pattern is accurately determined it is just a matter of time before fraud shifts to a different area exhibiting a totally different pattern. Unfortunately, detection models built with a two class classification approach cannot detect these new patterns. On the other hand, if we are able to develop a model that can identify the good non-fraud behaviors, then any pattern that does not categorize as good behavior can be considered outliers and subjected to further investigation. Keeping this idea in mind, we attempt to harness samples that capture good non-fraud patterns and use them to determine the outliers. We refer to the data points that exhibit good non-fraud patterns as consistent data points.
We consider data points that belong to the same cluster (or close proximity clusters) in different subspaces as being consistent. With this understanding we say that consistent data points have similar spatial arrangements in any subspace. This is especially true when the data points have features that are highly correlated. Based on this understanding, we attempt the problem of outlier detection through consistent data point detection. One way to get the most consistent data points from the sample data set is by applying an unsupervised technique like clustering. Since the spatial arrangements of consistent points do not change, they should form closed clusters. However, identifying the consistent points with one run of a clustering technique on the data set is not feasible as the cluster formation is very sensitive to the value of the number of clusters (k) that is selected. Also it is difficult to quantify the sense of consistency in the spatial arrangement with just one cluster arrangement. In order provide this information we rely on an ensemble method of clustering. We run the simple k-means clustering algorithm for different values of k where k ranges from 2 to K in steps p (depending on the data set). The K runs of k-means on the sample set will assign each data point to K clusters with C centroids associated with these clusters. A selection algorithm that captures the spatial arrangement is then applied on the centroids for each data point to determine consistency.

\begin{algorithm}{}\label{alg:algo}
\caption{$Consistent Data Selection Method$}
\begin{algorithmic}

\Require 
	\State $\Theta \leftarrow$ dataset
	\State AvgSimScore $\leftarrow n$ dimensional array
	\State $ \theta \leftarrow$ threshold for consistent data selection
	\State $C \leftarrow$ consistent dataset
	\State $I \leftarrow$ inconsistent dataset
\Procedure{}{}
	\State Run $k$-means on $\Theta$ for $K$ = \{$k_1$,$k_2$,$k_3$ .. $k_k$\}
	\For{$i=1...n$}
		\State $\mathbf{x_\emph{i}} \sim \Theta$ = $Set(\mathbf{C_1},\mathbf{C_2}..\mathbf{C_k})$, where $\mathbf{C_k}$ is a centroid of a cluster $\mathbf{x}$ belongs to
		\State AvgSimScore($i$) =$\sum_{l=1}^{l= \binom k2} cos(C_l,C_{l+1})/\binom k2$
		\If{AvgSimScore($i$) > $\theta$}
		\State $C \leftarrow C \cup \mathbf{x}$
		\Else
		\State $I \leftarrow I \cup \mathbf{x}$
		\EndIf
	\EndFor
\State  Train a one class classifier off $C$
\State  Label data points in $I$ to identify outliers

\EndProcedure
\end{algorithmic}
\label{alg:cds}
\end{algorithm}

\begin{figure}
\centering
\includegraphics[height=2in, width=3.5in]{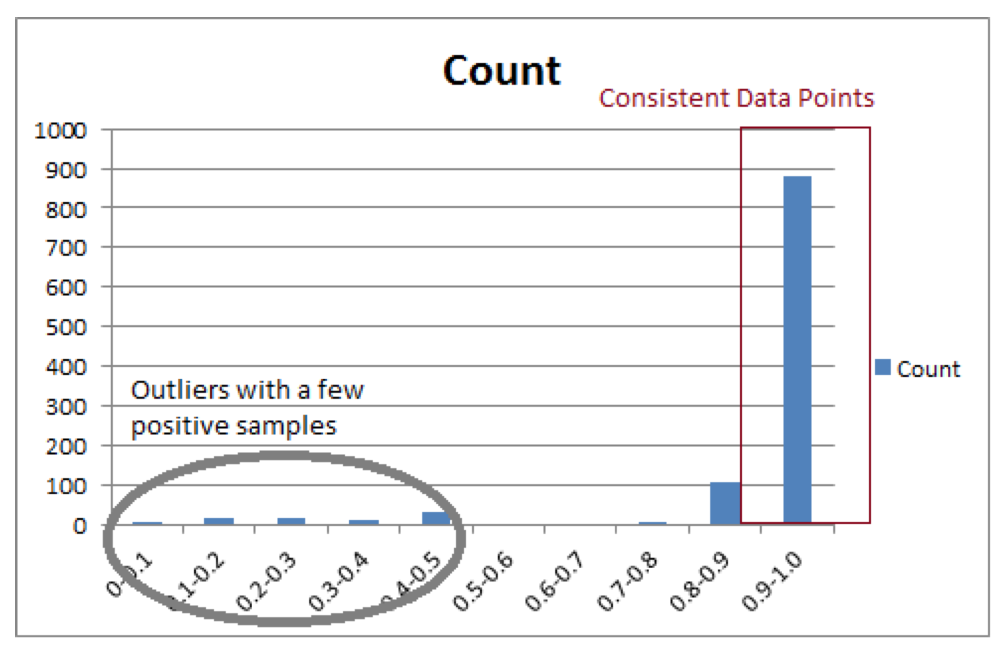}
\caption{Average Similarity score buckets for data points}
\label{fig:synth}
\vskip -6pt
\end{figure}

In order to give a visual sense of our hypothesis, we generated a synthetic data set with 1000:70 consistent (class 1) to outlier (class 0) data point ratio. We then clustered this data using k = $2$,5,10,100,500. A selection algorithm detailed in the next paragraph, is applied on the centroids of the clusters that each data point belongs to in the K cluster runs. To identify consistency in the spatial arrangement, we measure the vector similarity between the vector centroids to which each data point belongs to across the K cluster runs in the actual feature space. Each data point belongs to 5 clusters from 5 runs of k-means. The cluster centroids of these clusters are considered as vectors in the original feature space. Using the average of the cosine similarity scores between these centroid vectors, we verify to see if there is any centroid that does not fall in the same spatial arrangement. Data points in the sample set are ordered in the descending order of the centroid average similarity scores. The score buckets are as shown in Figure \ref{fig:synth}. Data points belonging to class 1 are prominent in the higher buckets while that of class 0 are clustered in the lower buckets, indicating that the spatial arrangement of the data points of class 1 is indeed consistent. The approach outlined on the synthetic data set helps to identify the consistent sample set that can be used in a one-class classifier to determine the outliers. The data points in the range of 0.8 - 0.1 are then used to train a one class SVM classifier using LIBSVM\cite{Chang:libsvm} with a polynomial kernel and the default parameters. This model is then applied on the remaining data points (score bucket 0 - 0.7) and the data points that cannot be classified as being similar to the training data are considered outliers. The results on the synthetic data set indicate the potential of the approach and justify our hypothesis that consistent data points do have similar spatial representation and identifying these points using an ensemble clustering approach is a good way to bootstrap samples needed to train an outlier detection model. 

\section{Consistency Detection}
\label{sec:cd}
Given a set of data points \{$\Theta$ = $\textbf{x}_1$,$\textbf{x}_2$,$\textbf{x}_3$ .. $\textbf{x}_n$\} that contain a relatively small sample of outliers  \{$O$ = $\textbf{o}_1$,$\textbf{o}_2$,$\textbf{o}_3$ .. $\textbf{o}_m$\} such that $m$ < < $n$, the goal is to effectively identify the outlier pool. Each data point can have multiple attributes and each attribute is denoted by $n_{ij}$ where $i$ is the data point index and $j$ is the attribute index. A consistent data point set is determined using an ensemble of $k$-means clusters, $K$ = \{$k_1$,$k_2$,$k_3$ .. $k_k$\}. The value of $k$ ranges from a small number to a large number where the largest run $k_k$ <= $n$, the sample size. Each data point, after running the suit of $k$-means clusters, will belong to $k$ clusters with $C$ centroids where $C$ = \{$\textbf{c}_1$,$\textbf{c}_2$, .. ,$\textbf{c}_k$\}. Based on our hypothesis, identifying consistent data points depend on identifying consistent clusters.

Since distance or density measure is ill defined in high dimensional space, we consider centroids as vectors in the actual feature/attribute space or in subspace and measure the similarity of the two centroid points based on vector similarity scores. For each data point, consistency is determined by estimating the average similarity of the centroids it belongs to.
\begin{displaymath}{AvgSimScore = \sum_{i=1}^{i= \binom k2} cos(C_i,C_{i+1})/\binom k2}\end{displaymath} where $cos$ is the cosine similarity metric and $C_i$ is the vector centroid of cluster $i$. If the average similarity score for a data point is very high (closer to 1) then the centroids are very close to each other, and the data point is considered to be consistent. Algorithm \ref{alg:cds} highlights all the steps of consistent data selection method.
\section{Experiments}
\label{sec:exp}
The motivation for this approach comes from trying to identify fraudulent consumers on an e-Commerce platform. On a data set that contains transactions for a given day, identifying fraudulent patterns is not easy. Each time the e-Commerce company introduces new consumer aided features or imposes restrictions on certain transactional behaviors, the good behavior pattern also exhibits a change and opens new doors and avenues for consumers to misuse and abuse the platform. It is not only important to detect the fraudulent transactions but also detect this shift in good behavior. Since there is no access to training data for either of the cases, we have to rely on unsupervised techniques that help bootstrap sampled to build further sophisticated learning models. Our algorithm shows tremendous potential in detecting transactions exhibiting good behavior and likely so, on applying it on a day's worth of transaction details, we obtained a good clean sample set. We also observed that the samples in the lower score buckets, had higher recall for fraudulent transactions. Due to the sensitivity of the data set, these results cannot be quantified in this paper. However in order to showcase the goodness of this technique, we apply this method
on three publicly available data sets from the UCI machine learning repository.

\subsection{Data Sets Descriptions and Preparation}
UCI machine learning repository provides access to many data sets. We selected three datasets - the Ionosphere dataset\footnote{\href{url}{http://archive.ics.uci.edu/ml/datasets/Ionosphere}}, the Arrhythmia dataset\footnote{\href{url}{http://archive.ics.uci.edu/ml/datasets/Arrhythmia}} and the Musk dataset\footnote{\href{url}{http://archive.ics.uci.edu/ml/datasets/Musk+(Version+2)}}. Though the primary task for these datasets is classification, we group certain class instances together and attempt this as an outlier detection task.
The selection of the outlier set can be done in many ways. One way is to select the outliers based on the feature distribution. Another way is to simply group classes that have instances less than certain threshold as outliers. In this work, for datasets with multiple classes, we group classes that occur in less than 5\% of the data set and consider them as outliers. Those with just two classes, class with lower number of instances are considered outliers.
The first set we tested with is the Ionosphere data set. It contains 351 instances with 34 attributes that help in determining if there exists some type of structure in the ionosphere. This data set was set up as a two class classification problem where the positive labels indicate that the signals were successful in identifying a structure in the ionosphere and negative labels indicate that the signals passed through the ionosphere and were not sensitive to the free electrons. There are 225 positive instances and 126 negative instances.
The Arrhythmia data set has 279 attributes corresponding to different measurements of physical and heart rate characteristics which are useful in diagnosing arrhythmia. The data set has 16 classes out of which 3 have zero entries. We group together classes 3,4,5,7,8,9,14 and 15 into the outlier set and group 1,2,6,10 and 16 into the consistent data set. With this grouping we now have and 386 consistent samples and 66 outlier samples. Another observation with this data set is that there exist many consistent samples with errors in the values, especially in recording the height and weight of a candidate sample. This may affect the results. However, we do not perform any sanity check on the data but use it as is in the algorithm run.
The third data set on which we ran our experiments is the Musk data set. This data set contains 6598 instances with 168 attributes. The attributes describe the molecular structure that helps categorize each instance into musk or non-musk category. There are 39 musk classes and 63 non- musk classes that are collapsed into two classes respectively. This gives us 1017 non-musk instances and 5581 musk instances, with the latter considered as outliers.

\section{Results}
\label{sec:result}
In this section we discuss the performance of our algorithm on each of the data sets, both in consistent data detection and one-class classification.
\subsection{Ionosphere Dataset}
\begin{figure}
\centering
\includegraphics[height=2.2in, width=3.5in]{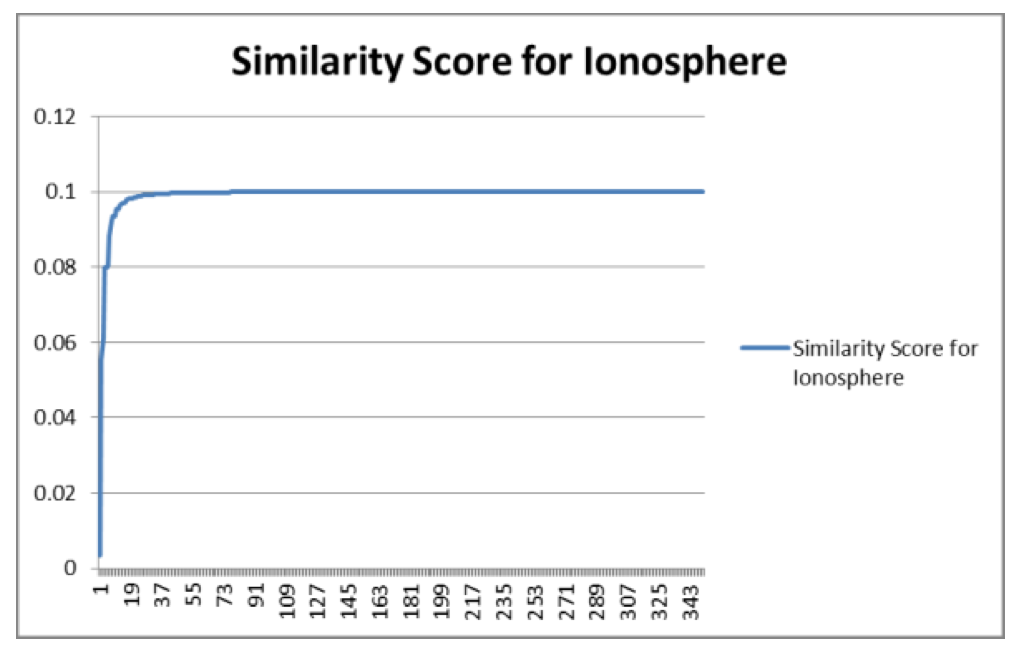}
\caption{Average similarity score of the centroids of the
Ionosphere data points. X-Axis is the count of data points available and Y-Axis is the score}
\label{fig:ionosphere}
\vskip -6pt
\end{figure}

The ionosphere data set contains a good mix of positive and negative samples. As mentioned before, in this experiment we consider the negative samples as outliers. In order to detect the consistent data points, the ensemble k-means clustering technique is applied on this data set with k =10,15,20,25,30,50,100,150,200,300. Each data point is assigned to 10 centroids. Figure \ref{fig:ionosphere} shows the average similarity score range for the centroids of each data point sorted in the descending order of the similarity score. 
Based on the above figure, we select all points with average score >= 0.1 threshold and consider them as consistent data points. The distribution of the samples is as shown in table~\ref{tab:a}. The distribution indicates that samples >= 0.1 score are indeed very consistent as majority of them have positive labels. Another observation is that the samples that are < 0.1 though have many positive samples are predominantly outliers. If this outlier set belonged to a sensitive task such as credit card fraud or e-Commerce fraud detection, then sending the entire set for manual review can work for high detection hit rate.

\begin{table}
\centering
%\caption{}
\begin{tabular}{|c|c|c|c|} \hline
Data Set &Data Split\protect\footnotemark&\# Positives & \# Negative\\ \hline
Consistent Pool & >=0.1 & 135&17\\ \hline
Inconsistent Pool & < 0.1 & 90&109\\ \hline

\hline\end{tabular}
\caption{Distribution of Ionosphere Dataset in the Split Buckets}\label{tab:a}
\end{table}
\footnotetext{based on average similarity score}
%\addtocounter{footnote}{1}

The second step is to extract the 109 outliers from the < 0.1 subset. This is done using the one-class classification approach. We explored two ways of determining the outliers. In the first approach we learn a multivariate Gaussian distribution over the consistent data set. For each data point in the inconsistent set, the distance from the Gaussian is calculated using the Mahalanobis distance measure\cite{mahalanobis1936generalized}. If this distance is greater than a predetermined threshold, then the data point is be flagged as an outlier. This method identified 60 true positive outliers but also identified 30 false positives. In our second approach we used the one class classifier with a polynomial kernel provided by LIBSVM. This classifier was trained on the consistent data set. When applied on the inconsistent data pool, we were able to successfully identify 87 outliers with only 17 false positives.

\subsection{Arrhythmia Dataset}
\begin{figure}
\centering
\includegraphics[height=2in, width=3.5in]{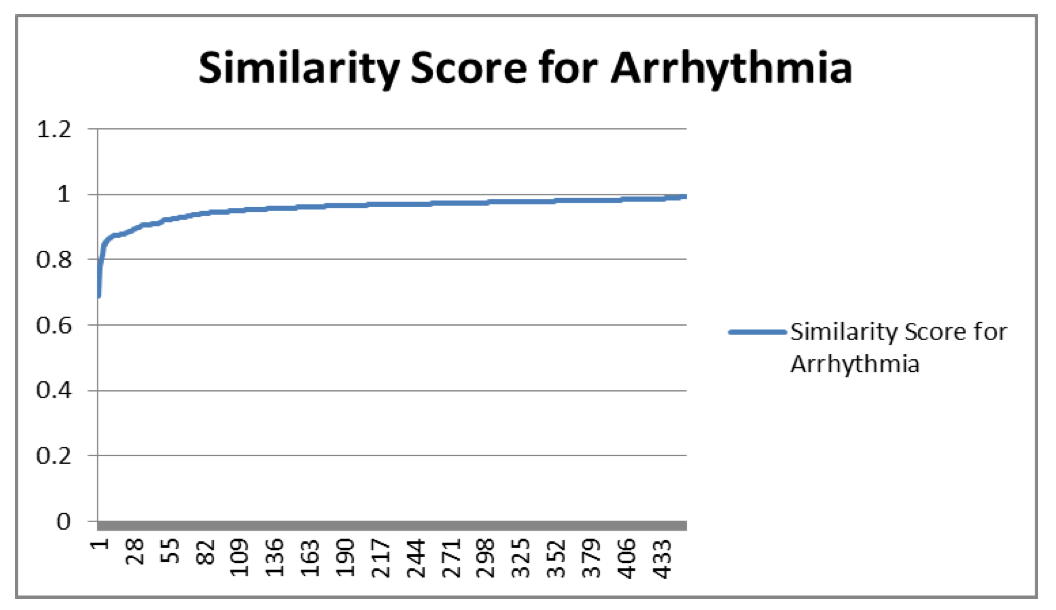}
\caption{Average similarity score of the centroids of the
Arrhythmia data points. X-Axis is the count of data points available and Y-Axis is the score}
\label{fig:arrhy}
\vskip -6pt
\end{figure}

The arrhythmia data set was set up to be a multi class classification problem. However, as mentioned before we have modified this data set for outlier detection. Several of the classes that are collapsed to fall in the outlier bucket have very few instances, and in some case as low as 2. Since our interpretation of outliers is simply based on the number of instances, the performance of our approach on the modified data set is not as good but yet shows promise. Also there are erroneous attribute entries in the positive samples. This adds to the challenge. The first step of generating consistent data set is executed with an ensemble of k=5,6,8,10,12,14,16,20,25,30,50,100. The average similarity scores on the centroids thus assigned are sorted in the descending order and the distribution plot is as shown in Figure \ref{fig:arrhy}.

We select a threshold of 0.95 based on the figure above and consider all the data samples with average similarity scores > 0.95 as consistent data points. The distribution of the samples in the pool after selection is as shown in table~\ref{tab:b}.

\begin{table}
\centering
%\caption{}
\begin{tabular}{|c|c|c|c|} \hline
Data Set &Data Split\protect\footnotemark&\# Positives & \# Negative\\ \hline
Consistent Pool & >0.95 & 275& 20\\ \hline
Inconsistent Pool & <= 0.95 & 111& 45\\ \hline

\hline\end{tabular}
\caption{Distribution of Arrhythmia Dataset in the Split Buckets}\label{tab:b}
\end{table}
%\addtocounter{footnote}{-2}
\footnotetext{based on average similarity score}
%\addtocounter{footnote}{1}

The consistent data set that was extracted shows promise in that the set has very few samples from the outlier set. However, there still is a large set of consistent samples in the inconsistent data pool. On careful examination of the actual negative samples in the inconsistent data pool, we notice that instances of classes (7,8,9,14 and 15) that are very sparse in the data set, the clear outliers, exist in this set. To further sanitize the inconsistent pool, we apply the consistent sample trained one class classifier on the in consistent set. All of the negative samples were classified as outliers.

However there were a considerable number of positives as well that were classified as outliers. Careful examination of some of these misclassified positive samples indicates that they are samples that have egregious attribute values.

\subsection{Musk Dataset}
The musk data set, though set up for a multi-class classification problem, can be easily formulated as a two class problem of identifying musk and non-musk samples, with non-musk samples being the outliers. Our first step of generating the consistent data set. We apply the ensemble k-means technique with k = 10, 15, 20, 25, 30, 50, 100, 150, 200, 300, 500, 800, 1000, 1500. On sorting the data points in the descending order of their average similarity score computed from the centroids they belong to in each of these clustering steps, we select a threshold of 0.4.

\begin{figure}
\centering
\includegraphics[height=2in, width=3.5in]{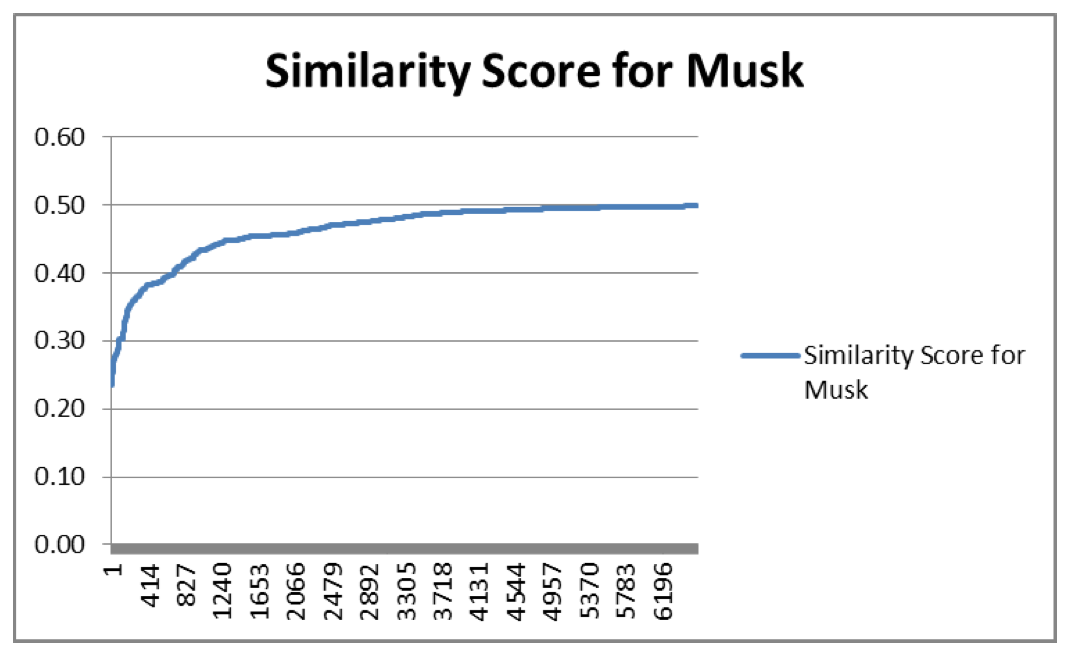}
\caption{Average similarity score of the centroids of the
Musk data points. X-Axis is the count of data points available and Y- Axis is the score}
\label{fig:musk}
\vskip -6pt
\end{figure}

\begin{table}
\centering
%\caption{Distribution of Musk Dataset in the Split Buckets}
\begin{tabular}{|c|c|c|c|} \hline
Data Set &Data Split\protect\footnotemark&\# Positives & \# Negative\\ \hline
Consistent Pool & >0.4 & 5538&0\\ \hline
Inconsistent Pool & <= 0.4 & 42&1017\\ \hline

\hline\end{tabular}
\caption{Distribution of Musk Dataset in the Split Buckets}\label{tab:c}
\end{table}
\footnotetext{based on average similarity score}
%\addtocounter{footnote}{1}

The distribution of the samples on splitting the data with the selected threshold is as shown in table~\ref{tab:c} and the efficacy of the method is clearly evident. Here the consistent pool accommodates most of the positive samples. The inconsistent pool has all of the negative samples (outliers) and very few positive samples. If this task were to be a very sensitive task of outlier detection, then subjecting the inconsistent data pool for manual review will be very useful. However, in order to extract only the outlier samples from the inconsistent pool, similar to what was done for the ionosphere and the arrhythmia data set, we train a one class classifier using a polynomial kernel and apply the learnt model to the inconsistent pool. This model was able to detect all of the outlier points and had only 15 positives being considered as outliers.

\section{Observations}
\label{sec:observation}
The algorithm to detect consistent data points is very sensitive to the variations of k that is used in the ensemble. If we do not use a good low value of k, then by the inherent goodness of the clustering algorithm in a given feature space, we will find distinct clusters that are outliers. This will affect the similarity measure calculation as the centroids of outliers will always be in a similar spatial arrangement and hence will be considered consistent. If a good range of k values is applied for clustering, then the benefit of our approach becomes clearly evident. One advantage of our method is that very easily we can bootstrap a very good quality inconsistent set that can be further processed to identify the outliers. We can also use the average similarity score as a ranking measure to select buckets of inconsistent sample sets that can be processed individually. The main benefit of our approach is that we do not require any apriori knowledge of data distributions that other statistical methods need, making it very discover outlier and consistent behavioral patterns.

\section{Conclusion}
\label{sec:conclusion}
In this paper we propose a method that shows tremendous potential in identifying outliers, especially in a scenario when training data is not available for either the consistent data set or the outlier set. The method focuses on extracting consistent data points from the set rather than extracting the outliers. The byproduct is a relatively clean set of outliers. In order to further improve the precision of outliers that are extracted we propose the use of a one-class classifier. This classifier is trained on the selected consistent data set and applied on rest of the sample pool. The goodness of this method is that it helps to quickly bootstrap samples in both the outlier set as well as the consistent data set, and these samples can be used to understand the behavioral patterns.

\bibliographystyle{abbrv}
\bibliography{sigproc}  % sigproc.bib is the name of the Bibliography in this case
\end{document}